# EMBRE: Entity-aware Masking for Biomedical Relation Extraction

Mingjie Li[1] and Karin Verspoor[1, *]

[1]School of Computing Technologies, RMIT University, Melbourne, AU

*Corresponding author: karin.verspoor@rmit.edu.au

**Abstract**
Information extraction techniques, including named entity recognition (NER) and relation extraction (RE), are crucial in many domains to support making sense of vast amounts of unstructured text data by identifying and connecting relevant information. Such techniques can assist researchers in extracting valuable insights. In this paper, we introduce the Entity-aware Masking for Biomedical Relation Extraction (EMBRE) method for biomedical relation extraction, as applied in the context of the BioRED challenge Task 1, in which human-annotated entities are provided as input. Specifically, we integrate entity knowledge into a deep neural network by pre-training the backbone model with an entity masking objective. We randomly mask named entities for each instance and let the model identify the masked entity along with its type. In this way, the model is capable of learning more specific knowledge and more robust representations. Then, we utilize the pre-trained model as our backbone to encode language representations and feed these representations into two multilayer perceptron (MLPs) to predict the logits for relation and novelty, respectively. The experimental results demonstrate that our proposed method can improve the performances of entity pair, relation and novelty extraction over our baseline.

**Introduction**
The BioRED challenge task (3) presented a rich biomedical relation extraction dataset, including 600 PubMed abstracts with gold standard annotations of 8 relation types connecting key entity types including drugs, diseases, genes, and chemicals. Additionally, the BioRED dataset includes an assignment of "Novel" or "No[t novel]" to each relation, indicating whether the relation comes from main content or background, similar to the focus/background entities of Jimeno Yepes and Verspoor (1) but extended to relations. BioRED provides a comprehensive testbed for researchers to build and evaluate their biomedical IE systems. Different from many existing RE datasets which focus on extracting relations from a single sentence, BioRED contains relations across multiple sentences, moving the research from sentence-level to document-level.

Along with the dataset, Luo et al. (3) also performed benchmarking experiments with several state-of-the-art methods for RE based on deep learning techniques. Specifically, they utilized BERT-GT (6) and PubMedBERT (5) as the backbone models to encode language representations. These encodings were then fed into classifiers to predict specific relation types. The experimental results demonstrate that using PubMedBERT – which is pre-trained on literature from PubMed – is more effective than the BERT (2) language model for biomedical RE. The PubMedBERT token vocabulary recognizes more words from the BioRED corpus than general BERT. In addition, it encodes more biomedical knowledge and generates robust and generalizable token embeddings for this task. Building on this dataset, several deep learning-based RE systems have been proposed. For example, BioREx (4) provides a data-centric approach that improves the benchmark performances by leveraging a rich dataset combining five RE datasets with different kinds of annotations. This approach injects diverse, synergistic, biomedical knowledge into the model.

In this paper, we explore a method we dub EMBRE (Entity-aware Masking for Biomedical Relation Extraction) to incorporate knowledge into the benchmark model without the need for a large, heterogeneous corpus. PubMedBERT is trained on literature from PubMed with the masked language modelling (LM) objective. In a RE task, a critical step is to recognize and label the key named entities involved in the relation. Although the PubMedBERT tokenizer can identify the relevant vocabulary tokens, the random token-level masking strategy does not guarantee that these entity embeddings are effectively captured during pre-training. Therefore, we propose a new masked pre-training strategy, which randomly masks full named entities within each instance from BioRED, comprising of title and abstract. During this pre-training, we give the model the identifier and concept type of each masked token, as these are the two key elements of information for each entity. This enriched information in the input assists the model in encoding understanding of named entities into robust token embeddings. Then we employ two multilayer perceptron (MLPs) to predict the logits for the relation classification and novelty identification tasks, respectively. We fine-tune the whole model in a multi-task learning paradigm, addressing both of these tasks together. The experimental results on the BioRED Task1 test set demonstrate that the EMBRE masking strategy can significantly improve the performance of a benchmark model on both relation extraction and novelty identification.

## Methods
### Named Entity Masking Pre-training Strategy EMBRE
In an attempt to bolster the understanding and recognition of pairs of related named entities, our proposed methodology hinges on the development of a new masking pre-training strategy, as illustrated in Figure 1. In the proposed strategy, named entities are randomly masked within each instance, where an instance encompasses both the title and abstract for a given PMID in the BioRED dataset.

We denote the named entity pairs by $P = \{p_1, p_2, \ldots, p_n | p_i = (e_{src}, e_{tgt})\}$. We then utilize a masking function $M(.)$ to randomly mask entities with [MASK] tokens. Masking all the entities during pretraining cannot improve the entity representation learning process. Notably, we did not mask all the paired entities so that the model retains the capability of embedding entities. However, we did mask all entity mentions corresponding to the same identifier. For multi-word entities, we mask all the words, averaging the learned representations to obtain the entity representations. During this pre-training, PubMedBERT is tasked with identifying both the identifier and concept type for each masked token, as these two components encapsulate key information for each entity. The incorporation of named entity masking into pre-training facilitates PubMedBERT to achieve a deeper understanding of named entities, efficiently encoding them into robust token embeddings. We call this EMBRE: Entity-aware Masking for Biomedical Relation Extraction.

### Multi-task Fine-tuning
The architecture of our model for RE is illustrated in Figure 1. The overall model is fine-tuned in a multi-task learning paradigm, ensuring that the model generalizes well on both relation extraction and novelty identification tasks concurrently. After the pre-training phase, given the robust token embeddings from our pre-training strategy, two independent MLPs are designed to predict the logits for relation and novelty, respectively. The objective functions to be minimized during fine-tuning can be written as:

$$L = -\lambda_1 \sum_{c_r} y_r \log(\widehat{y_r}) - \lambda_2 \sum_{c_n} y_n \log(\widehat{y_n})$$

Where $\lambda_1$ and $\lambda_2$ are the weights to balance the loss, $c_r$ and $c_n$ refer to the classes of relation types and novelty categories, $y$ and $\hat{y}$ are the ground truth and prediction of each case. In this study, we empirically set $\lambda_1$ to 1 and $\lambda_2$ to 2. We utilize `itertools` from Python to generate all the entity pair candidates, including all positive and negative samples, for training and inference.

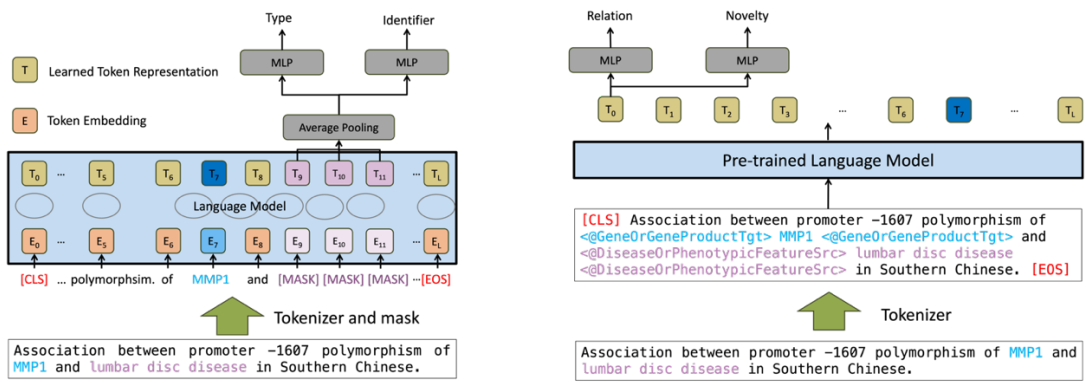

**Figure 1. (left)** Illustration of our proposed entity masking pre-training strategy. Example from PMID: 18439317. **(right)** Architecture of our full model for predicting the relation extraction and novelty identification from given paired entities. The subjective and objective entities are in blue and purple texts, respectively.

**Experimental Setting**
We implement our model by using Python 3.9, PyTorch 1.13 with CUDA 11.6. We first pre-train the PubMedBERT loaded from HuggingFace for 15 epochs with BioRED corpus. The masking threshold is set to 0.2. Then we load the pre-trained language model and fine-tune the whole network for 15 epochs and utilize the checkpoint achieving the best performances on validation set for testing. Both pre-training and fine-tuning are conducted on 5 NVIDIA 3080 Ti GPUs with batch sizes of 128. An Adam optimizer, with a learning rate of 1e-5, is applied. For each case, we combine the title and abstract, and add a specific tag before and after the source and target entities to represent their type, like the official setting.

**Results**
For Run 1, we utilize the base pre-trained PubMedBERT as the backbone to encode the input language into representations and two MLPs for the prediction. In Run 2, we utilize our proposed model based on entity-aware masking. We report the detailed performances for each task in Table 1. Adopting the F1-score as the main metric for comparison, our method significantly improves the performance of vanilla PubMedBERT. The baseline model yields a Precision of approximately 0.402, Recall of 0.115, and an F1-score of 0.179 overall for the combined tasks of named entity recognition (NER), relation extraction (RE) plus novelty detextion (ND) classification of the related pair. The EMBRE method attains a Precision of about 0.234, Recall of 0.217, and an F1-

score of 0.225; thereby increasing Recall at a cost of Precision, while F1 benefits overall. This trend is consistent in relation to the overall average and median results across all submissions to the shared task; the baseline Recall performance of our models is low and while EMBRE contributes to a substantial increase, the corresponding decrease in Precision limits overall performance.

**Table 1.** Results of Runs 1 (base PubMedBERT) and 2 (base + EMBRE) and average (Avg.) and median (Med.) of all the submissions on BioRED test set.

|  |  | Precision | Recall | F1 |  | Precision | Recall | F1 |
|---|---|---|---|---|---|---|---|---|
| Run 1 | NER | **0.7236** | 0.2369 | 0.3560 | NER | **0.5236** | 0.1765 | 0.2640 |
| Run 2 |  | 0.5273 | **0.4926** | **0.5094** | + RE | 0.3335 | **0.3160** | **0.3245** |
| BioRED Avg. |  | *0.6922* | *0.6860* | *0.6703* |  | *0.4901* | *0.4839* | *0.4774* |
| BioRED Med. |  | *0.7793* | *0.6965* | *0.7356* |  | *0.5164* | *0.5479* | *0.5317* |
| Run 1 | NER | **0.5515** | 0.1539 | 0.2407 | NER | **0.4017** | 0.1151 | 0.1789 |
| Run 2 | +ND | 0.3724 | **0.3412** | **0.3561** | +RE | 0.2341 | **0.2173** | **0.2254** |
| BioRED Avg. |  | *0.5092* | *0.5002* | *0.4923* | +ND | *0.3615* | *0.3573* | *0.3522* |
| BioRED Med. |  | *0.5297* | *0.6042* | *0.5645* |  | *0.4161* | *0.3988* | *0.4073* |

## Discussion

In the exploration of biomedical text mining, distinct performance disparities emerge between the baseline and our proposed model. While the baseline, leveraging pre-trained PubMedBERT, demonstrates high precision at the sacrifice of recall, our EMBRE method strikes a better balance between the two, enhancing the breadth and accuracy of relationship extraction. The named entity masking pre-training strategy coupled with the multi-task learning paradigm bolster the model's proficiency in comprehending and extracting entity relationships and novelty. Compared with ERNIE (7), which incorporates structured external knowledge and dynamic masking based on context, our method specifically masks named entities and only integrates annotations provided directly in BioRED during fine-tuning. This means it relies only on task-specific data. Furthermore, we ask the model to predict both identifier and type which can inject informative content into our language model. In this paper, while we improve the baseline model with EMBRE, we do not make other notable changes in our approach, leading to overall performance below the median. Further experimentation is needed to explore EMBRE in conjunction with other changes such as tweaking our assumed input representations.

## Conclusion

In this paper, we introduce the EMBRE entity masking pretraining method to address document-level named entity relation extraction, in the BioRED corpus. Specifically, we propose a named entity masking pre-training strategy to inject entity-specific knowledge into the model. This strategy facilitates entity understanding and enhances token embeddings. After fine-tuning using a multi-task learning paradigm, the results on BioRED test set demonstrate EMBRE significantly improves the baseline.

## Acknowledgements

This work was supported with funding from the Australian Research Council Linkage Project LP160101469 and Elsevier BV.

# References

1. Jimeno Yepes, Antonio and Karin Verspoor. "Distinguishing between focus and background entities in biomedical corpora using discourse structure and transformers". In *Proc 13th Intl Wkshp on Health Text Mining and Information Analysis (LOUHI)*, pp. 35–40. (2022).
2. Devlin, Jacob, Ming-Wei Chang, Kenton Lee and Kristina Toutanova. "BERT: Pre-training of deep bidirectional transformers for language understanding." In *Proc NAACL-HLT*, vol. 1, p. 2. (2019).
3. Luo, Ling, Po-Ting Lai, Chih-Hsuan Wei, Cecilia N. Arighi, and Zhiyong Lu. "BioRED: a rich biomedical relation extraction dataset." *Briefings in Bioinformatics* 23, no. 5 (2022): bbac282.
4. Lai, Po-Ting, Chih-Hsuan Wei, Ling Luo, Qingyu Chen, and Zhiyong Lu. "BioREx: Improving Biomedical Relation Extraction by Leveraging Heterogeneous Datasets." *arXiv preprint arXiv:2306.11189* (2023).
5. Gu, Yu, Robert Tinn, Hao Cheng, Michael Lucas, Naoto Usuyama, Xiaodong Liu, Tristan Naumann, Jianfeng Gao, and Hoifung Poon. "Domain-specific language model pretraining for biomedical natural language processing." *ACM Transactions on Computing for Healthcare (HEALTH)* 3, no. 1 (2021): 1-23.
6. Lai, Po-Ting, and Zhiyong Lu. "BERT-GT: cross-sentence n-ary relation extraction with BERT and Graph Transformer." *Bioinformatics* 36, no. 24 (2020): 5678-5685.
7. Zhang, Zhengyan, Xu Han, Zhiyuan Liu, Xin Jiang, Maosong Sun, and Qun Liu. "ERNIE: Enhanced Language Representation with Informative Entities." In *Proc 57th conf of the Assoc for Computational Linguistics*, pp. 1441-1451. (2019).